\documentclass[a4paper,conference]{IEEEtran}
\usepackage{cite}
\usepackage{amsthm, amssymb, amsmath}
\usepackage{graphicx}
\usepackage{float}
\usepackage{rotating}
\usepackage{subfigure}
\usepackage[justification=centering]{caption}
\usepackage[lined,ruled,linesnumbered,commentsnumbered]{algorithm2e}
\usepackage{cleveref}
\usepackage{booktabs}
\usepackage{amsmath,breqn}
\usepackage{multirow}
\usepackage{amsmath}

\usepackage{pdflscape}

\usepackage{color, colortbl}

\newtheorem{definition}{Definition}

\newtheorem{remark}{Remark}

\usepackage{amsmath}
\usepackage{array}
\usepackage{url}
\IEEEoverridecommandlockouts

\begin{document}

\bibliographystyle{IEEEtran}
%
% paper title
% Titles are generally capitalized except for words such as a, an, and, as,
% at, but, by, for, in, nor, of, on, or, the, to and up, which are usually
% not capitalized unless they are the first or last word of the title.
% Linebreaks \\ can be used within to get better formatting as desired.
% Do not put math or special symbols in the title.

\title{Solving Dynamic Multi-objective Optimization Problems Using Incremental Support Vector Machine}

\author{Weizhen~HU,~\IEEEmembership{}
	  Min JIANG$^*$,~\IEEEmembership{Senior~Member,~IEEE,}\hfil\break	
	  Xing Gao,~\IEEEmembership{}\hfil\break\\
	Kay Chen TAN,~\IEEEmembership{Fellow,~IEEE,}\hfil\break~
   and Yiu-ming Cheung,~\IEEEmembership{Fellow,~IEEE}\hfil\break
% <-this % stops a space
	\IEEEcompsocitemizethanks{\IEEEcompsocthanksitem W.Hu and M. JIANG are with the Department of Cognitive Science, Xiamen University, China, Fujian, 361005. X. Gao is with the Software School of Xiamen University. Min JIANG is the corresponding author and email: minjiang@xmu.edu.cn.
		% note need leading \protect in front of \\ to get a newline within \thanks as
		% \\ is fragile and will error, could use \hfil\break instead.
		\IEEEcompsocthanksitem KC TAN is with the Department of Computer Science, City University of Hong Kong.
       \IEEEcompsocthanksitem Yiu-ming Cheung is with the Department of Computer Science, Hong Kong Baptist University.}% <-this % stops an unwanted space
	
}

% make the title area
\IEEEoverridecommandlockouts
\IEEEpubid{\makebox[\columnwidth]{978-1-5386-4362-4/18/\$31.00~\copyright2018
IEEE \hfill} \hspace{\columnsep}\makebox[\columnwidth]{ }}
\maketitle
\begin{abstract}
The main feature of the Dynamic Multi-objective Optimization Problems (DMOPs) is that optimization objective functions will change with times or environments. One of the promising approaches for solving the DMOPs is reusing the obtained Pareto optimal set (POS) to train prediction models via machine learning approaches. In this paper, we train an Incremental Support Vector Machine (ISVM) classifier with the past POS, and then the solutions of the DMOP we want to solve at the next moment are filtered through the trained ISVM classifier. A high-quality initial population will be generated by the ISVM classifier, and a variety of different types of population-based dynamic multi-objective optimization algorithms can benefit from the population. To verify this idea, we incorporate the proposed approach into three evolutionary algorithms, the multi-objective particle swarm optimization(MOPSO), nondominated sorting genetic algorithm II (NSGA-II), and the regularity Model-based multi-objective estimation of distribution algorithm(RE-MEDA). We employ experiments to test these algorithms, and experimental results show the effectiveness.
\end{abstract}

% Note that keywords are not normally used for peerreview papers.

% For peer review papers, you can put extra information on the cover
% page as needed:
% \ifCLASSOPTIONpeerreview
% \begin{center} \bfseries EDICS Category: 3-BBND \end{center}
% \fi
%
% For peerreview papers, this IEEEtran command inserts a page break and
% creates the second title. It will be ignored for other modes.
\IEEEpeerreviewmaketitle

\section{Introduction}
One of the essential characteristics of Dynamic Multi-objective Optimization Problems (DMOPs) \cite{Farina_2004} is that the optimization functions will change with environments or times, and it is of great significance for many practical applications \cite{Cruz_2010}. Therefore, efficiently solving DMOPs has become an important direction in the field of evolutionary computation community \cite{Nguyen_2012}. For example, when designing a missile, the defense department should consider its range, precision, weight and fuel consumption. This is a four-objective optimization problem, and these optimization objectives will vary depending on environments and times. However, Most of the existing methods do not perform well when dealing with such problems. The main difficulty is how to quickly track the changing Pareto optimal front (POF).

In recent years, great progress has been made in this field, and many different algorithms have been proposed. Among those algorithms, the methods based on prediction have attracted special attention, and the basic idea of the method is to solve DMOPs by reusing ``experience'' effectively.  For example, Muruganantham \textit{et al.} \cite{Muruganantham_2016} proposed a Kalman Filter prediction based DMOEA(MOEA/D-KF) that utilizes a prediction model based on Kalman Filter based on the multiobjective EA with Decomposition.

In this paper, we argue that Incremental Support Vector Machine (ISVM) \cite{Laskov2006Incremental}\cite{tuvsar2007differential} can be used to train a prediction model, and the model helps any kind of population-based DMOPs algorithms by generating a high-quality initial population. This method has unique advantages in dealing with DMOPs, because it directly establishes connections between different POS, without requiring more computational resources to repeatedly train SVM classifiers.

The contribution of this research is to propose an algorithm which combining evolutionary multi-objective optimization algorithm with the ISVM technique. This combination has the following two advantages. First, the proposed design can improve the search accuracy in the way of reusing past experience. Second, this method trains the SVM classifier in an online manner, and then obtains the initial population for the next time, which can make more efficient use of computing resources.

The rest of this paper is organized as follows: In Section \ref{sec:Preliminaries-and-Related}, we will introduce some basic concepts of dynamic optimization problems, Incremental Support Vector Machine ( ISVM ) and related work. In Section \ref{sec:ISVM-DMOEA}, we will propose the Incremental Support Vector Machine based Dynamic Multi-Objective Evolutionary optimization Algorithm, ISVM-DMOEA. In Section \ref{experiments} we firstly introduce the evaluation criteria, test examples and comparative methods, and then experimental results are analyzed. In Section \ref{Conclusion}, we conclude the main work of this research and the future research direction are discussed.

\section{Preliminaries and Related Works}
\label{sec:Preliminaries-and-Related}

\subsection{Concepts of Dynamic Multi-objective Optimization}
For dynamic multi-objective optimization problems, its optimization functions often vary with times or environments. Mathematically, a DMOP can be described as:
$$\textbf{minimize}~f\left(x,t\right)=\left<f_{1}\left(x,t\right),f_{2}\left(x,t\right),...,f_{m}\left(x,t\right)\right>$$
$$s.t.\ x\in\Omega, $$
where $t$ represents time or environment index and $x=\left<x_{1},x_{2},\ \ldots,\ x_{n}\right>$ is the decision vector. $m$ is the number of objectives, $f_{i}\left(x,t\right):\Omega\ \rightarrow\ \mathbb{R}\ \left(i=1,\ \ldots,\ M\right)$ which is the objective space.

\begin{definition}{\emph{[Dynamic Decision Vector Domination]}}
	At a particular moment $t$ , a decision vector $x_{b}$ is Pareto dominated by another vector $x_{a}$ , it can be expressed as $x_{a}\succ_{t}x_{b}$, if and only if\emph{:}
	\begin{equation}
	\begin{cases}
	\forall i=1,\ldots,m, & f_{i}(x_{a},t)\leq f_{i}(x_{b},t)\\
	\exists i=1,\ldots,m, & f_{i}(x_{a},t)<f_{i}(x_{b},t)
	\end{cases}.
	\end{equation}
\end{definition}

\begin{definition}{\emph{[Dynamic Pareto-optimal Set]} }
	At time t, if and only if there is no other decision vector $x$ can Pareto dominate decision vector $\ensuremath{x^{*}}$, the $\ensuremath{x^{*}}$ is called Pareto optimal solution. All Pareto optimal solutions make up Dynamic Pareto-optimal Set(DPOS) at time $t$, that is\emph{:}
	$$DPOS=\left\{ x^{*}|\not \exists x,~x\succ_{t}x^{*}\right\}.$$
\end{definition}

\begin{definition}{\emph{[Dynamic Pareto-optimal Front]}}
	 The Dynamic Pareto-Optimal Front (DPOF) is the set of corresponding objective vectors of the Pareto-optimal Solution at time $t$.	
	$$DPOF=\left\{ f\left(x^{*},t\right)|x^{*}\in DPOS\right\}.$$
\end{definition}

\subsection{Incremental Support Vector Machines}
Support Vector Machine ( SVM )\cite{boser1992training} was proposed in 1964 , developed rapidly after the 1990s and spawned a series of improved and extended algorithms.  The SVM becomes a well-known learning method used for classification problems \cite{Osuna1997An}, and it is a generalized linear classifier for binary classification of data in supervised learning. Its decision boundary is the maximum-margin hyperplane that is solved for learning samples. Incremental Support Vector Machine (ISVM) is the combination of Online learning technology and the SVM. Incremental technology has been developed to facilitate bulk SVM learning \cite{Cauwenberghs2000Incremental} on very large data sets and has been widely used in the SVM community.

Given the training data $\{x_1, \cdots, x_N\}$ and learning objective $\{y_1, \cdots, y_N\}$ in the classification problem, where each sample of training data contains multiple features and thus constitutes a feature space. The learning objective is a binary variable $y_i \in \{1,-1\}$ representing negative class and positive class. A Support Vector Machine (SVM) is a discriminative classifier formally defined by a separating hyperplane, which can be described as:
$$\mathbf{w^T} \times x+b=0, $$ where $\mathbf{w}\in \mathbb{R}^t$ and $b\in \mathbb{R}$.

In linear inseparable problems, the use of SVM with hard margins will generate classification errors, so a new optimization problem can be constructed by introducing loss function on the basis of maximizing margins. The optimization problem of soft margin SVM is shown as follows \cite{Diehl2003SVM}:
\begin{eqnarray}
&\mathbf{Minmize_{(w,b)}}~\frac{1}{2}||\mathbf{w}||^2 + l\cdot \sum_{i=1}^{N}\varepsilon_i \nonumber\\
& \mathbf{ subj.~to:~} y_i(\mathbf{w}\times x_i+b)\geq 1-\varepsilon_i, i = i \cdots N.
\label{SVM}
\end{eqnarray}
Where $l$ is a constant and the second term of Equation (\ref{SVM}) provides an upper bound for the error in the training data, and the first term  makes maximum margin of separation between classes.

when learning nonlinear SVMs, to simplify matters and then this quadratic program is typically expressed in its dual form:

\begin{equation}
\min_{0\leq \alpha _i\leq l} W=\frac{1}{2} \sum_{i,j=1}^{N} \alpha_iQ_{ij}\alpha _j-\sum_{i=1}^{N} \alpha _i +b\sum_{i=1}^{N} y_i\alpha _i \nonumber\\
\label{SVM-1}
\end{equation}
Where ${Q_{ij}} = {y_i}{y_j}K({{\bf{x}}_i},{{\bf{x}}_j})$,$K({{\bf{x}}_i},{{\bf{x}}_j})= \varphi \left ( x_i \right )\cdot \varphi \left ( x_j \right ) $ is the given kernel function which to implicitly map into a higher (possibly infinite) dimensional feature space.

The Karush-Kuhn-Tucker (KKT) condition uniquely defines the solution of dual parameters$\left\{{\alpha,b}\right\}$ , and minimizes the form (2):
\[\begin{array}{l}
{g_i} = \frac{{\partial W}}{{\partial {\alpha _i}}} = \sum\limits_{j = 1}^N {{Q_{ij}}{\alpha _j} + {y_i}b - 1 = \left\{ \begin{array}{l}
 > 0{\rm{  }}\quad { \alpha _i} = 0\\
 = 0{\rm{  }}\quad  0 < {\alpha _i} < l\\
 < 0{\rm{  }}\quad { \alpha _i} = l
\end{array} \right.} \\
h = \frac{{\partial W}}{{\partial b}} = \sum\limits_{j = 1}^N {{y_j}{\alpha _j}}  = 0
\end{array}\]

The KKT conditions partition the training data into three parts:

${g_c} = 0$, the set S of margin support vectors;

${g_c} \le 0$, the set E of error support vectors;

${g_c} > 0$, the set R of the remaining vectors.

When we increment the unlearned examples into the solution, our goal will be to keep all previously seen training data in KKT conditions simultaneously\cite{Laskov2006Incremental}.

\subsection{Related Works}
The DMOPs field has made great progress, and existing algorithms can be generally divided into the following categories: Diversity based Approaches, parallel approaches, Memory based Approaches and Change prediction based Approaches .

An important work of diversity based approaches is the Dynamic NSGA-II (DNSGA-II) proposed by \cite{Deb2007Dynamic} in 2006. Deb \textit{et al.} extended NSGA-II to deal with DMOPs by introducing diversity in each change detection. If the target or constraint violation value has changed, the problem is considered to have changed. Then, all outdated solutions (that is, reassessment). This process allows for the use of changing objectives and constraint functions to evaluate both offspring and parent solutions. In \cite{Chen2009Using}, Chen proposed to maintain genetic diversity by regard it as an additional objective when solving multi-objective optimization. They presented the Individual Diversity Evolutionary Method (IDEM) to add a useful selection pressure for optimizing PS and maintaining diversity \cite{Chen2009Using}. The results show that the algorithm can converge to the optimal solution effectively and track the change of PFs while maintaining the diversity of solution sets.

Parallel EAs utilize several subpopulations that evolve simultaneously under different processors and then communicate some informations in a structured network \cite{Al2002Parallel}. In \cite{C2009A}, Camara \textit{et al.} proposed a method to apply parallel single pre-genetic algorithm (PSFGA) to dynamic environment. PSFGA is a master-slave architecture algorithm, which divides the population into subpopulations which performs a certain amount of generations and preserve only non-dominant solutions, and then the main process joins all the solutions to a new population.

The memory-based approach uses additional memory to implicitly or explicitly store useful information from past generations to guide future searches. It has been proved that when the optimal solution returns to the previous position repeatedly or the environment changes periodically, this algorithm will help save computing time and bias search process, thus becoming very efficient. In \cite{Azzouz2015A}, Azzouz \textit{et al.} proposed an adaptive hybrid population management strategy, which is based on a technology that can measure the severity of environmental changes, then according to the technology it can adjust memory, local search (LS) and the number of random solutions.

When the behavior of dynamic problems follows a certain trend, a prediction model can be used. In 2014, the author defined a new prediction model \cite{li2014dynamic} to solve DMOPs with Translational optimal POS (DMOP-tps). Dmop-tps is a specific type of DMOP in which POS is converted periodically over time. When the environment changes, the strategy proposed by Deb \textit{et al.} \cite{Deb2007Dynamic} is used to detect the changes. Then, the population is re-initialized according to the properties of DMOP. Muruganantham \textit{et al.} \cite{Muruganantham_2016} proposed a Kalman Filter prediction based DMOEA(MOEA/D-KF) based on the multiobjective EA with Decomposition. When the change of environment is detected, kalman filter is applied to the whole population, which leads the search to the new pareto optimal solution in the decision space. The influence of the severity and frequency of change on the performance of the this algorithms is studied.

\section{INCREMENTAL SUPPORT VECTOR MACHINE BASED DYNAMIC MULTI-OBJECTIVE OPTIMIZATION ALGORITHM }
\label{sec:ISVM-DMOEA}
A dynamic multi-objective optimization problem is based on the fact that different environments follow different possible distributions, and these are not independent, but interrelated. For the dynamic multi-objective optimization problem, it is easy to get the solution. However, it is difficult for us to judge whether this solution is good or bad, but the solution already obtained should contain useful information, which can be used to predict POS at the next moment. So in this article, we turn the decision problem into a classification problem. In other words, we are constantly training the ISVM classifier through the information we have obtained, that is, the POS of past moments.

At the same time, a good initial population is the key to solving the dynamic multi-objective problem. A good initial population not only speeds up the solution, but also improves the quality and accuracy of the solution. By training the SVM classifier in an online way, we can use this continuously improved classifier to filter out the solutions, and then generate the initial population at the next moment.

In this article, we propose a incremental support vector machine (ISVM) based dynamic optimization algorithm.
The details of the proposed algorithm is descried in Algorithm.  \ref{alg:SMOTEandISVM-DMOEA}.

This algorithm is mainly divided into two modules. At the initial moment, we combine the solution (positive example) in POS obtained by using multi-objective evolutionary algorithm and the randomly generate some solutions of the problem (non-POS, negative example) into training samples, and then obtain a SVM classifier $SC_S$. When the environment changes, we regard the changes as the incremental learning process of SVM classifier for the change process of dynamic multi-objective optimization problems. At each change moment, we consider the obtained POS and non-POS as samples to update the parameters of $SC_S$ classifier. At the same time, using $SC_S$ classifier, we can classify the solutions in the next moment into two categories, ``good'' and ``bad''. And the good solutions is reserved as the initial population of the next moment. Finally, we can get the $SC_S$ classification model with the best performance.

\label{sec:SMOTEandISVM-DMOEA}

\begin{algorithm}
	\caption{ISVM-DMOEA:Incremental Support Vector Machines based Dynamic Multi-objective Evolutionary Algorithm}
	\label{alg:SMOTEandISVM-DMOEA}
	\KwIn{ The Dynamic Multi-objective Optimaztion Function $F(X)$;  }
	\KwOut{$POSs$: the POSs of $F(X)$;}
	Randomly initiate a Population the $Pop_0$\;
	$POS_0$ =DMOEA($Pop_0$)\;
	$POS_s$	=$POS_0$\;
    Train a SVM classifier $SC_S$ by using $Pg \in POS_{0}$ and $Ng \notin POS_{0}$\;
	Randomly generate solutions $\{xy_1,\cdots,xy_{p}\}$ of the function $F(X)_{1}$\;
    \If{$xy_i$ pass the recognition of the SVM $SC_S$ }{Put $xy_i$ into $Pop_{1}$}
    $POS_1$=DMOEA($Pop_1$)\;
    $POS_t$	=$POS_1$\;
	\For{$t = 1$ to $n$}{
        $PSAMPLES_t$=$POS_t$\;
		Train $SC_S$ by using $Pg \in PSAMPLES_{t}$ and $Ng \in NSAMPLES_{t}$\;
		Randomly generate solutions $\{xy_1,\cdots,xy_{p}\}$ of the function $F(X)_{t+1}$\;
		\If{$xy_i$ pass the recognition of the SVM $SC_S$ }{Put $xy_i$ into $Pop_{t+1}$}
		$POS_{t+1}$ = DMOEA($Pop_{t+1}$)\;
		$POSs$ = $POSs \cup POS_{t+1}$ \;		
	}
	\Return $POSs$\;
\end{algorithm}

\begin{remark}
	In this research, we only used the NSGA-II, the MOPSO and the RM-MEDA algorithms to obtain POS at next time, but in fact, any population-based algorithm can use our proposed method to achieve performance improvements.
\end{remark}

\begin{remark}
  The $PSAMPLES_t$ at particular moment are obtained by $POS_t$ while $NSAMPLES_{t}$ are randomly generate\cite{Guo2016Learning},and they have the same number.
\end{remark}

\begin{remark}
   At the initial moment, we train a SVM classifier $SC_S$ by using $Pg \in POS_{0}$ and $Ng \notin POS_{0}$\ , and when t change from 1 to n, we still train the same $SC_S$ using $Pg \in PSAMPLES_{t}$ and $Ng \in NSAMPLES_{t}$.
\end{remark}
\section{Empirical Study}
\label{experiments}
In general, our proposed algorithm is applicable to any kind of population-based optimization algorithm. In our experiment, we chose three representative experiments to prove our method. The first multiobjective optimization algorithm is based on particle swarm optimization \cite{Coello2002MOPSO}, it was called MOPSO for short. The second one is the NSGA-II and it is a multiobjective genetic algorithm that applies nondominated sorting and crowding distance. The third one is a distribution estimation algorithm \cite{zhang2008rm} based on global statistical information to construct a probability model, We simply refer to it as the RM-MEDA. Obviously, these three algorithms belong to different categories, and they were not originally designed for dynamic optimization. But they are well developed, so we can improve our level of persuasion and confidence in the technologies we propose. The three corresponding algorithms with incremental SVM are called ISVM-MOPSO, ISVM-NSGA-II, and ISVM-RMMEDA respectively. It is worth noting that parameters, such as population size, iteration number are all the same. In other words, for the three groups of algorithms, we did not deliberately adjust the experimental parameters for getting better performances.
\subsection{Performance Metrics, Testing Functions and Settings}
In this study, when comparing with other competitive algorithms, we use inversed generational distance (IGD) and its variants as performance indicators to evaluate the quality of solutions obtained by different algorithms.

$1)$The inverted generational distance (IGD) \cite{sierra2005improving} is an index to measure the distance between the real Pareto Optimal Front expressed by $Q^*$, and the approximate Pareto Optimal Front obtained by the algorithm, which we use $Q$ to represent.Then the definition of the IGD can be described as：
	\begin{equation}
	\label{def:igd}
	\mathrm{IGD}(Q^*, Q, C) = \frac{\sum_{p^* \in Q^*}\min_{p\in Q}\left\|p^*-p\right\|}{\left|Q^*\right|}.
	\end{equation}
	
It is worth noting that the definition of IGD in this paper is slightly different from the original definition. The main difference is the parameter $C$ in Equation (\ref{def:igd}), which we call the configuration of the benchmark functions. The configurations used in this experiment are shown in Table \ref{table:settings}. IGD compares the ideal POF with the POF obtained by other algorithms. If the distance between P and P* is closer, the smaller the value of IGD, the higher the performance of the algorithm is to some extent.

One variant of the IGD, called MIGD, can also be used to evaluate dynamic multiobjective optimization algorithms \cite{Muruganantham_2015,Muruganantham_2016}. MIGD is the average value of IGD values at a certain time step in each run, described as:
	\begin{equation}
	\mathrm{MIGD}(Q^*, Q, C) = \frac{1}{|T|}\sum_{t \in T}\mathrm{IGD}(Q_t^{*}, Q_{t}, C),
	\end{equation}
Where $T$ represents the set of all moments. At time t, $Q_t^{*}$ represents the point set of the ideal POF, and $Q^{t}$ represents the approximate POF obtained by the algorithm. In addition, we also hope to evaluate these algorithms in a dynamic environment, so we have defined a new indicator DMIGD based on MIGD. The definition of DMIGD is as follows:
	
	\begin{equation}
	\mathrm{DMIGD}(Q^*, Q, C) = \frac{1}{|E|}\sum_{C \in E}\mathrm{MIGD}(Q_t^{*}, Q_{t}, C),
	\end{equation}
	
Where $|E|$ is the number of different environments experienced. We used eight different environmental configurations to conduct our experiments. It is worth noting that DMIGD allows us to evaluate the dynamic multi-objective optimization algorithm from a high-level perspective, and MIGD only considers the dynamics in an environment, so it is obviously different from MIGD.

\subsection{Test Instances and Experimental Settings}
In this research, we take the IEEE CEC 2015 benchmark problems set as test functions and the problem set has eleven testing functions. Details of the functions definitions are given in \cite{helbig2015benchmark}. In the definitions, the decision variables are $x=(x_1,\ldots,x_n)$ and $t = \frac{1}{n_t}\left\lfloor \frac{\tau_T}{\tau_t} \right\rfloor$, where $n_t, \tau_T$, and $\tau_t$ are the severity of change, maximum number of iterations, and frequency of change respectively. Table \ref{table:settings} describes the different combinations of $n_t$, $\tau_t$, and $\tau_T$ used in our experiments. Please note that, for each $n_t$-$\tau_T$ combination, there will be $\frac{\tau_T}{\tau_t}$ environment changes. In other words, in all of our experiments,  every test function requires 20 environmental transformations for each pair of $n_t$-$\tau_T$  combination.

\begin{table}[!htbp]
	\centering
	\caption{Environment Settings}
	\label{table:settings}
	\begin{tabular}{cccc}
		\toprule
		& \textbf{$n_t$} & \textbf{$T_t$} & \textbf{$T_T$} \\
		\midrule
		\textbf{C1} & 10    & 5     & 100 \\
		\textbf{C2} & 10    & 10    & 200 \\
		\textbf{C3} & 10    & 25    & 500 \\
		\textbf{C4} & 10    & 50    & 1000 \\
		\textbf{C5} & 1     & 10    & 200 \\
		\textbf{C6} & 1     & 50    & 1000\\
		\textbf{C7} & 20    & 10    & 200 \\
		\textbf{C8} & 20    & 50    & 1000 \\
		\bottomrule
	\end{tabular}%	
\end{table}

The POFs of the  testing functions have different shapes and each function belongs to a certain DMOPs type. For example, the POF of dMOP3 and DIMP2 is convex while the POF of HE2 is discontinuous. and for FDA5,the spread of POF solutions changes over time. Table \ref{table:instances} describes the types of the testing functions. Type I means POS changes, but POF does not; Type II implies that  when POS changes, POF changes accordingly; Type III indicates that POF has changed, but POS has not.
\begin{table}[!htbp]
	\caption{Characteristic of the test functions}
	\label{table:instances}
	\centering
	\begin{tabular}{|l|c|c|c|}
		\hline
		Name & \begin{tabular}{c}Decision \\Variable \\Dimension\end{tabular} & Objectives & DMOP Type \\\hline
		FDA4          & 12 & 3 & TYPE I   \\\hline
		FDA5          & 12 & 3 & TYPE II  \\\hline
		DIMP2         & 10 & 2 & TYPE I   \\\hline
		dMOP2         & 10 & 2 & TYPE II  \\\hline
		HE7                   & 10 & 2 & TYPE III \\\hline
		HE9                   & 10 & 2 & TYPE III \\\hline
	\end{tabular}
	
\end{table}

As depicted in Table \ref{table:instances}, the dimensions of the decision variables are 10 and 12, In all experiments, the population size was set to 200. As mentioned earlier, for each environment configuration, we changed each benchmark function 20 times, and in every change, we let the entire population perform 50 iterations in the evolutionary algorithm. For the incremental Support Vector Machines, we set the kernel type is Gaussian kernel and kernel scale is obtained by the Grid Search method \cite{fine2002incremental}.

\subsection{Experimental Results}
In this study, we conducted two different kinds of experiments. In the first kind of experiment, we incorporate the proposed approach into the development of three well-known evolutionary algorithms, multiojective particle swarm optimization (MOPSO), nondominated sorting genetic algorithm II (NSGA-II), and the regularity model-based multiobjective estimation of distribution algorithm (RM-MEDA). We employ six  benchmark functions to test these algorithms. The experimental results are recored in Table. \ref{tab:MIGD1}. In almost all experiments, the proposed method has been significantly improved compared with the original method.

In the second kind of experiment, we compared the ISVM-RM-MEDA with some chosen algorithms, and the results are recorded in Table. \ref{tab:MIGD2}. The MBN-EDA \cite{karshenas2014multiobjective} is an estimation algorithm, in which the dependence between decision variables and target variables is obtained by using a multidimensional Bayesian network. MOEA\textbackslash{}D-KF was presented in \cite{Muruganantham_2016} and this idea for solving DMOPs is to predict the decision space by Kalman filter.
 SVM-NSGA-II was proposed in \cite{8377567}  and it is a support vector machine based dynamic MOPs algorithm.

For incremental SVM, a major iteration corresponds to contain a new example. Therefore, the computational complexity estimates must be multiplied by the number of training instances learned so far. The actual runtime depends on the balance between arithmetic operations and memory access in a small iteration. In our proposed method, when adding new samples incrementally, we can use the SVM classifier that has been obtained in the past moment to simplify the computation cost of searching the solution of the next quadratic programming. By the way, for samples that are not partitioned into the set $s$ of margin support vectors, there is no need to carry out new iterations and update core parameters, which is the key module for computation cost in incremental SVM usage.

\begin{table*}[htbp]
	\centering
	\caption{The MIGD Values between NSGA-II, ISVM- NSGA-II, MOPSO, ISVM-MOPSO, RM-MEDA and ISVM- RM-MEDA}
	\begin{tabular}{l|cc|cc|cc}

		\toprule
		\textbf{DMIGD} & \textbf{NSGA-II} & {\textbf{ISVM-NSGAII}} & \textbf{MOPSO} & {\textbf{ISVM-MOPSO}} & \textbf{RM-MEDA} & {\textbf{ISVM-RM-MEDA}} \\
		\midrule
		FDA4  & 0.2783  & \textbf{0.1839 } & 0.0812  & \textbf{0.0731  }& 0.0684  & \textbf{0.0639}  \\
		FDA5  & 0.3580  & \textbf{0.2604}  & 0.2510  & \textbf{0.1586 } & 0.2345  & \textbf{0.1157}  \\
		DIMP2 & 3.9818  & \textbf{2.5154 } & 2.6386  & \textbf{2.3410 } & 4.9631  & 5.2002  \\
		DMOP2 & 0.6590  & \textbf{0.0922 } & 0.2984  & \textbf{0.0881 } & 4.5942  & \textbf{4.5228}  \\
		HE7   & 0.0946  & \textbf{0.0534 } & 0.0636  & \textbf{0.0561 } & 0.0428  & \textbf{0.0342 } \\
		HE9   & 0.2945  & \textbf{0.2473}  & 0.2502  & \textbf{0.2450}  & 0.2581  & \textbf{0.2337 } \\
		\bottomrule
	\end{tabular}%
	\label{tab:MIGD1}%
\end{table*}%

\begin{table*}[htbp]
	\centering
	\caption{The MIGD Values between MBN-EDA, MOE/D-KF,SVM-NSGA-II,ISVM-RM-MEDA}
	\begin{tabular}{lcccc}

		\toprule
		\textbf{DMIGD} & \textbf{MBN-EDA} & {\textbf{MOE/D-KF}} & {\textbf{SVM\newline{}-NSGA-II}} & {\textbf{ISVM-RMMEDA}}\\
		\midrule
		FDA4  & 0.4300  & 0.1913  & 0.2166  & \textbf{0.0639 } \\
		FDA5  & 0.5100  & 0.4963  & 0.2910  & \textbf{0.1157 } \\
		DIMP2 & 6.9700  & 22.9536  & \textbf{2.8463}  & 5.2002  \\
		DMOP2 & 1.4000  & 3.0619  &\textbf{0.1356}  & 4.5228  \\
		HE7   & 0.2100  & 0.2365  & 0.0728  & \textbf{0.0342}  \\
		HE9   & 0.3600  & 0.4108  & 0.2497  & \textbf{0.2337}  \\
		\bottomrule
	\end{tabular}%
	\label{tab:MIGD2}%
\end{table*}%

\section{Conclusion}
\label{Conclusion}
Evolutionary algorithms can solve dynamic multi-objective optimization problems more effectively through good initial populations, but getting a good initial population is a difficult and resource-consuming problem.  In order to avoid wasting valuable computing resources on repeated computing, our basic idea is that incremental training support vector machine classifiers can be trained by using the POS have been obtained at different times, and when changes occur, the parameter of SVM classifier will be updated in an online manner, then the trained classifier will filter the solution at the next moment, and the good individuals will be selected as the initial population. This initial population can help any kind of population-based algorithms to solve the dynamic multi-objective optimization problem quickly and more accurately.

This research is just a new starting point, after which we will learn how to combine incremental support vector machines with transfer learning\cite{7349204eee} or imbalanced learning techniques \cite{8100935}. On the other hand, we also want to study use deep learning methods \cite{jiang2014improving} to automatically generate better samples to solve the real world problems, such as motion generation of Multi-Legged Robot\cite{jiang2017motioneee} and dynamic path planning \cite{jiang2012fuzzy}.
% use section* for acknowledgment
\section*{Acknowledgment}
This work was supported by the National Natural Science Foundation of China (No.61673328).

% Can use something like this to put references on a page
% by themselves when using endfloat and the captionsoff option.
\ifCLASSOPTIONcaptionsoff
  \newpage
\fi

% trigger a \newpage just before the given reference
% number - used to balance the columns on the last page
% adjust value as needed - may need to be readjusted if
% the document is modified later
%\IEEEtriggeratref{8}
% The "triggered" command can be changed if desired:
%\IEEEtriggercmd{\enlargethispage{-5in}}

% references section

% can use a bibliography generated by BibTeX as a .bbl file
% BibTeX documentation can be easily obtained at:
% http://mirror.ctan.org/biblio/bibtex/contrib/doc/
% The IEEEtran BibTeX style support page is at:
% http://www.michaelshell.org/tex/ieeetran/bibtex/
%\bibliographystyle{IEEEtran}
% argument is your BibTeX string definitions and bibliography database(s)
%\bibliography{IEEEabrv,../bib/paper}
%
% <OR> manually copy in the resultant .bbl file
% set second argument of \begin to the number of references
% (used to reserve space for the reference number labels box)
%\begin{thebibliography}{1}
\bibliography{mybibtex}

%\end{thebibliography}

% biography section
%
% If you have an EPS/PDF photo (graphicx package needed) extra braces are
% needed around the contents of the optional argument to biography to prevent
% the LaTeX parser from getting confused when it sees the complicated
% \includegraphics command within an optional argument. (You could create
% your own custom macro containing the \includegraphics command to make things
% simpler here.)
%\begin{IEEEbiography}[{\includegraphics[width=1in,height=1.25in,clip,keepaspectratio]{mshell}}]{Michael Shell}
% or if you just want to reserve a space for a photo:

% that's all folks
\end{document}